\pdfoutput=1

\documentclass[11pt,letterpaper]{article}
\usepackage{emnlp2016}
\usepackage{times}
\usepackage{latexsym}
\usepackage{amsmath}
\usepackage[hang,small,bf]{caption}
\usepackage[subrefformat=parens]{subcaption}
\usepackage[pdftex]{graphicx}
\usepackage{bm}
\usepackage{amssymb}
\usepackage{amsfonts}

\emnlpfinalcopy

\def\emnlppaperid{1059}

\usepackage{xcolor}

\title{Controlling Output Length in Neural Encoder-Decoders}

\author{Yuta Kikuchi$^1$ \\ \texttt{\small kikuchi@lr.pi.titech.ac.jp}
\And Graham Neubig$^2$\thanks{\ \ This work was done when the author was at the
Nara Institute of Science and Technology.} \\ \texttt{\small gneubig@cs.cmu.edu}
\And Ryohei Sasano$^1$ \\ \texttt{\small sasano@pi.titech.ac.jp}
\AND
Hiroya Takamura$^1$ \\ \texttt{\small takamura@pi.titech.ac.jp}
\And Manabu Okumura$^1$ \\ \texttt{\small oku@pi.titecjh.ac.jp}
\AND \vspace{-7mm} \\
$^1$Tokyo Institute of Technology, Japan \\ $^2$Carnegie Mellon University, USA}

\date{}

\def\modelA{$fixLen$}
\def\modelB{$fixRng$}
\def\modelC{$LenEmb$}
\def\modelD{$LenInit$}

\newcommand{\encfh}[1]{\overrightarrow{\bm{h}}_{#1}}
\newcommand{\encfc}[1]{\overrightarrow{\bm{c}}_{#1}}
\newcommand{\encbh}[1]{ \overleftarrow{\bm{h}}_{#1}}
\newcommand{\encbc}[1]{ \overleftarrow{\bm{c}}_{#1}}
\newcommand{\dech}[1]{\bm{s}_{#1}}
\newcommand{\decc}[1]{\bm{m}_{#1}}

\newcommand{\bhline}[1]{\noalign{\hrule height #1}}

\begin{document}
\interfootnotelinepenalty=10000

\maketitle

\begin{abstract}
  Neural encoder-decoder models have shown great success in many sequence
generation tasks.
  However, previous work has not investigated situations in which we would
  like to control the length of encoder-decoder outputs.
  This capability is crucial for applications such as text summarization,
  in which we have to generate concise summaries with a desired length.
  In this paper, we propose methods for controlling the output sequence length
  for neural encoder-decoder models: two decoding-based methods and two
  learning-based methods.\footnote{Available at https://github.com/kiyukuta/lencon.}
  Results show that our learning-based methods have the capability to control
  length without degrading summary quality in a summarization task.
\end{abstract}

\section{Introduction}
\label{sec:intro}

Since its first use for machine translation \cite{kalchbrenner13,cho14,sutskever14}, the
encoder-decoder approach has demonstrated great success in many other sequence
generation tasks including image caption generation \cite{vinyals15a,xu15a},
parsing \cite{vinyals15b}, dialogue response generation \cite{li16a,serban16}
and sentence summarization \cite{rush15,chopra16}.
In particular, in this paper we focus on sentence summarization, which as its name suggests, 
consists of generating shorter versions of sentences for applications such as
document summarization \cite{nenkova11} or headline generation \cite{dorr03}.
Recently, Rush et al. \shortcite{rush15} automatically constructed large
training data for sentence summarization, and this has led to the rapid development of neural sentence
summarization (NSS) or neural headline generation (NHG) models.
There are already many studies that address this
task \cite{nallapati16,ayana16,ranzato15,lopyrev15,gulcehre16,gu16,chopra16}.

One of the essential properties that text summarization systems should have is
the ability to generate a summary with the desired length.
Desired lengths of summaries strongly depends on the scene of use, 
such as the granularity of information the user wants to understand, 
or the monitor size of the device the user has.
The length also depends on the amount of information 
contained in the given source document.
Hence, in the traditional setting of text summarization, 
both the source document and the desired length of the summary will be
given as input to a summarization system.
However, methods for controlling the output sequence length of encoder-decoder
models have not been investigated yet, despite their importance in these
settings.

In this paper, we propose and investigate four methods for controlling the
output sequence length for neural encoder-decoder models.
The former two methods are decoding-based; they receive the
desired length during the decoding process, and the training process is the same as
standard encoder-decoder models.
The latter two methods are learning-based;
we modify the network architecture to receive the desired length as input.

In experiments, we show that the learning-based methods outperform the
decoding-based methods for long (such as 50 or 75 byte) summaries.
We also find that despite this additional length-control capability, the
proposed methods remain competitive to existing methods on standard settings of
the DUC2004 shared task-1.

\section{Background}

\subsection{Related Work}

Text summarization is one of the oldest fields of study in natural language processing,
and many summarization methods have focused specifically on sentence compression or headline generation.
Traditional approaches to this task focus on word deletion using rule-based
\cite{dorr03,zajic04} or statistical \cite{woodsend10,galanis10,filippova08,filippova13,filippova15}
methods.
There are also several studies of abstractive sentence summarization using syntactic
transduction \cite{cohn08,napoles11} or taking a phrase-based
statistical machine translation approach \cite{banko00,wubben12,cohn13}.

Recent work has adopted techniques such as encoder-decoder \cite{kalchbrenner13,sutskever14,cho14}
and attentional \cite{bahdanau14,luong15} neural network models from the field of machine translation,
and tailored them to the sentence summarization task.
Rush et al. \shortcite{rush15} were the first to pose
sentence summarization as a new target task for neural
sequence-to-sequence learning.
Several studies have used this task as one of the benchmarks of their neural sequence transduction methods
\cite{ranzato15,lopyrev15,ayana16}.
Some studies address the other important phenomena frequently occurred in
human-written summaries, such as copying from the source document \cite{gu16,gulcehre16}.
Nallapati et al. \shortcite{nallapati16} investigate a way to solve many
important problems capturing keywords, or inputting multiple sentences.

Neural encoder-decoders can also be viewed as statistical language models conditioned
on the target sentence context.
Rosenfeld et al. \shortcite{rosenfeld01} have proposed whole-sentence language models
that can consider features such as sentence length.
However, as described in the introduction,
to our knowledge, explicitly controlling length of output sequences in neural language models
or encoder-decoders has not been investigated.

Finally, there are some studies to modify the output
sequence according some meta information such as the dialogue act~\cite{wen15},
user personality \cite{li16b}, or politeness~\cite{sennrich16}.
However, these studies have not focused on length, the topic of this paper.

\begin{figure*}[t]
  \begin{minipage}[b]{0.33\linewidth}
    \centering
    \includegraphics[keepaspectratio, scale=0.42]{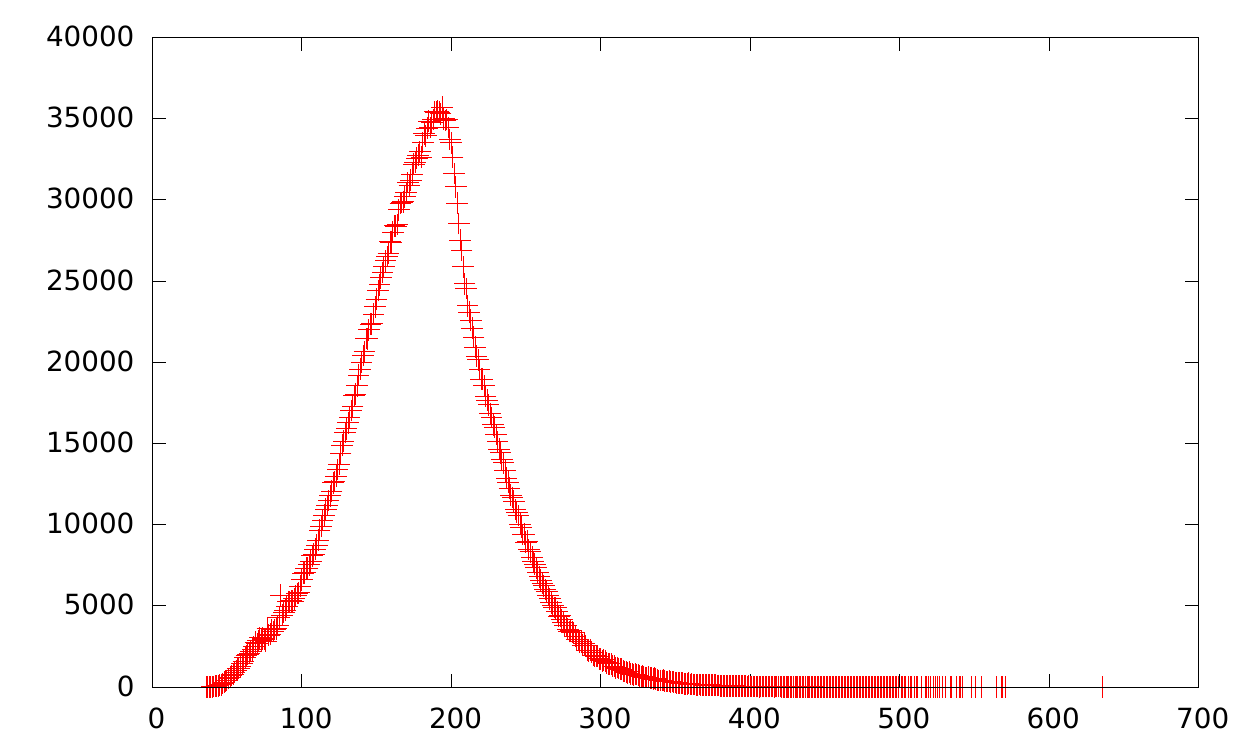}
    \subcaption{\footnotesize first sentence (181.87)}\label{fig:article_hist}
  \end{minipage}
  \begin{minipage}[b]{0.33\linewidth}
    \centering
    \includegraphics[keepaspectratio, scale=0.42]{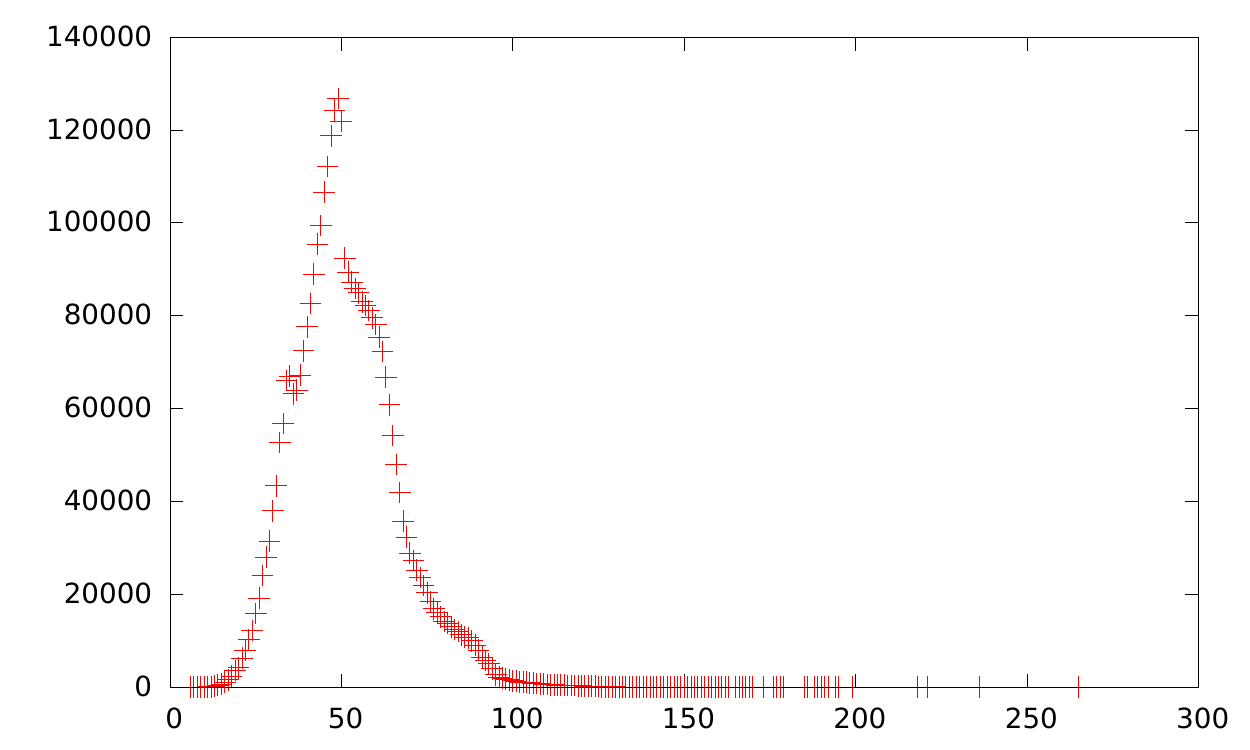}
    \subcaption{\footnotesize article headline (51.38)}\label{fig:title_hist}
  \end{minipage}
  \begin{minipage}[b]{0.33\linewidth}
    \centering
    \includegraphics[keepaspectratio, scale=0.42]{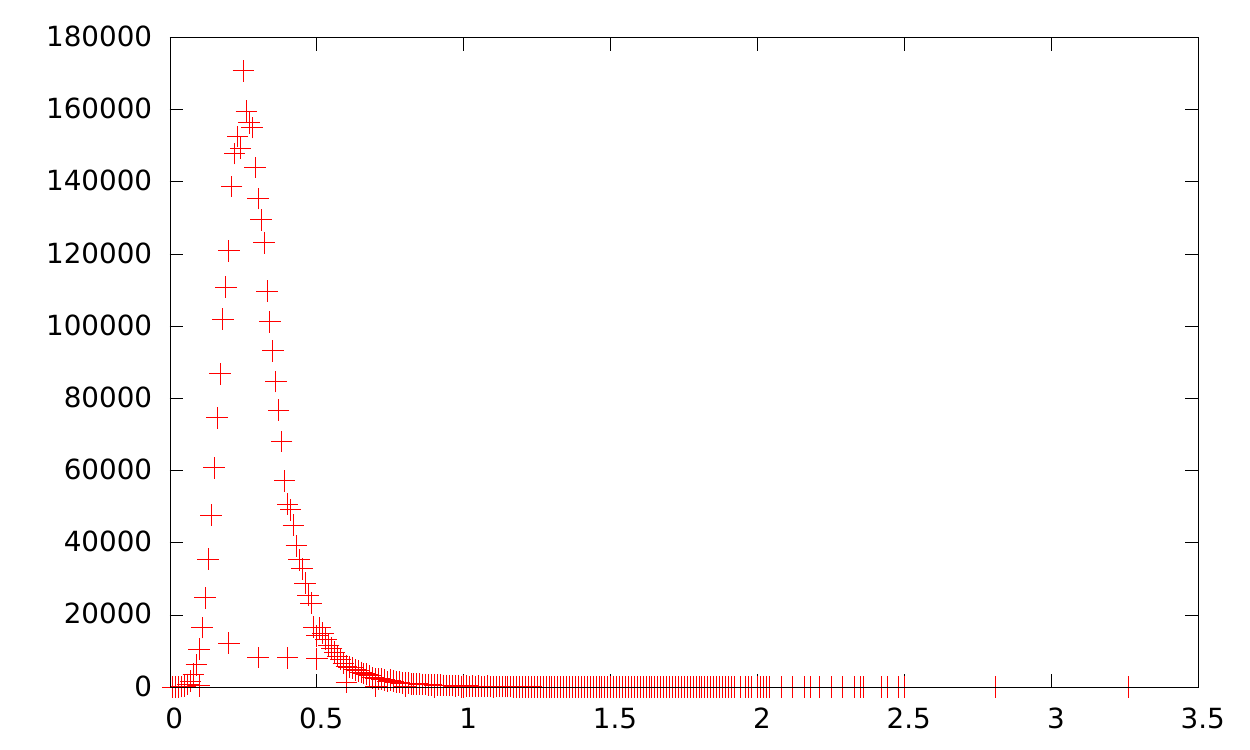}
    \subcaption{\footnotesize ratio (0.30)}\label{fig:ratio_hist}
  \end{minipage}
  \caption{Histograms of first sentence length, headline length, and their ratio in
Annotated Gigaword English Gigaword corpus.  Bracketed values in each subcaption are averages.}
\label{fig:agiga_hist}
\end{figure*}

\subsection{Importance of Controlling Output Length}
\label{sec:focus_length}

As we already mentioned in Section \ref{sec:intro},
the most standard setting in text summarization is to input both the source
document and the desired length of the summary to a summarization system.
Summarization systems thus must be able to generate summaries of various
lengths.
Obviously, this property is also essential for summarization methods based on
neural encoder-decoder models.

Since an encoder-decoder model is a completely data-driven approach, the output
sequence length depends on the training data that the model is trained on.
For example, we use sentence-summary pairs extracted from the Annotated
English Gigaword corpus as training data~\cite{rush15}, and the average length
of human-written summary is 51.38 bytes.
Figure \ref{fig:agiga_hist} shows the statistics of the corpus.
When we train a standard encoder-decoder model and perform the
standard beam search decoding on the corpus, the average length of its output
sequence is 38.02 byte.

However, there are other situations where we want summaries with other lengths.
For example, DUC2004 is a shared task where the maximum length of summaries is
set to 75 bytes, and summarization systems would benefit from generating sentences
up to this length limit.

While recent NSS models themselves cannot control their output length,
Rush et al. \shortcite{rush15} and others following use 
an ad-hoc method, in which the system is inhibited from generating the
end-of-sentence (EOS) tag by assigning a score of $-\infty$ to the tag and
generating a fixed number of words\footnote{According to the published code (https://github.com/facebook/NAMAS),
the default number of words is set to 15, which is too long for the DUC2004 setting.
The average number of words of human summaries in the evaluation set is 10.43.},
and finally the output summaries are truncated to 75 bytes.
Ideally, the models should be able to change the output sequence depending on
the given output length, and to output the EOS tag at the appropriate time point
in a natural manner.

\begin{figure}[t]
  \centering
  \includegraphics[keepaspectratio, scale=0.45]{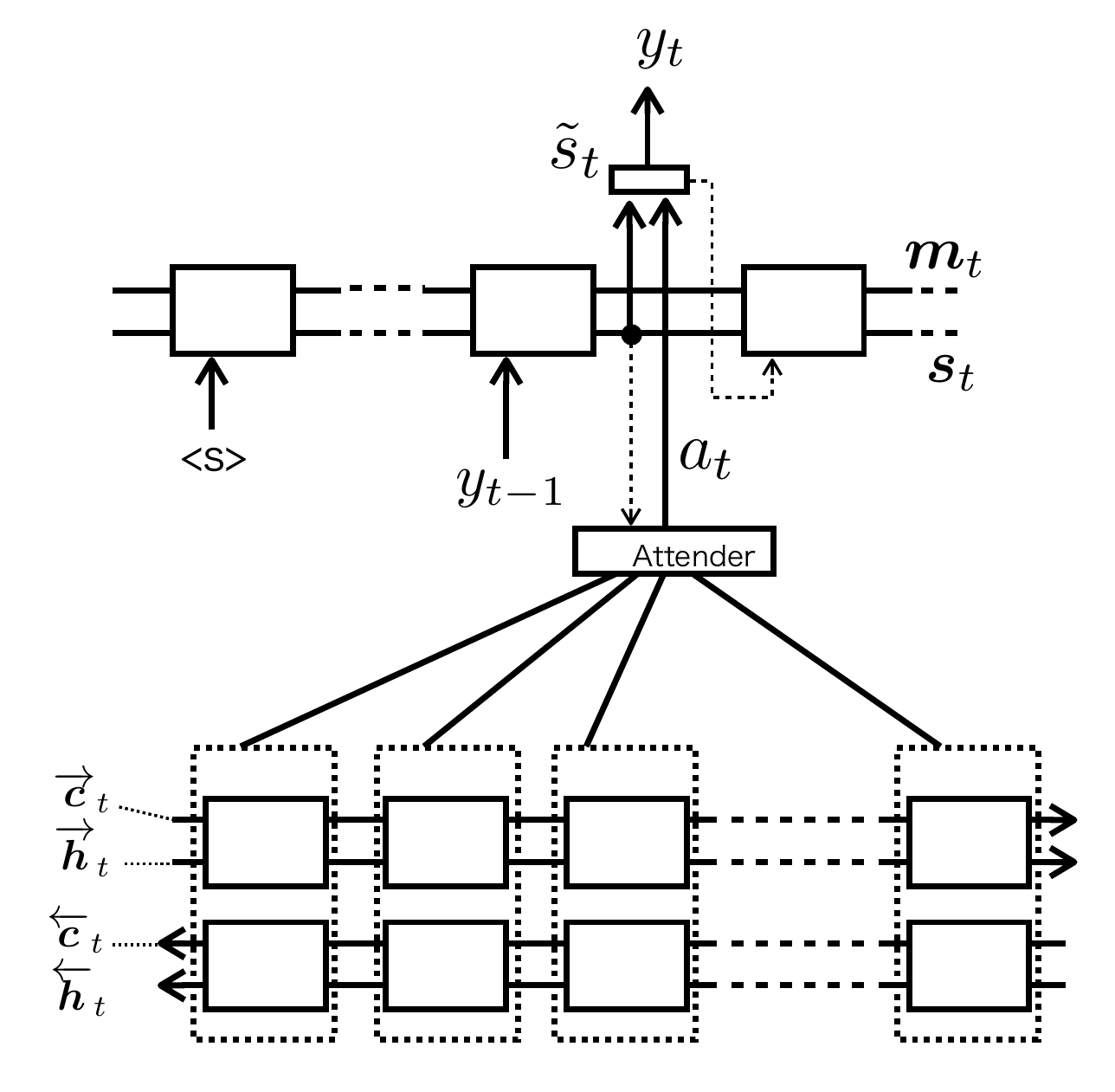}
  \caption{The encoder-decoder architecture we used as a base model in this
paper.}
  \label{fig:encdec}
\end{figure}

\section{Network Architecture: Encoder-Decoder with Attention}
\label{sec:encdec}

In this section, we describe the model architecture used for our experiments:
an encoder-decoder consisting of bi-directional RNNs and an attention mechanism.
Figure \ref{fig:encdec} shows the architecture of the model.

Suppose that the source sentence is represented as a sequence of words $\bm{x} = (x_1,
x_2, x_3, ..., x_N)$.
For a given source sentence, the summarizer generates a shortened version of the 
input (i.e. $N > M$), as summary sentence  $\bm{y} = (y_1, y_2, y_3, ..., y_M)$. 
The model estimates conditional probability $p(\bm{y}|\bm{x})$ using parameters trained on
large training data consisting of sentence-summary pairs.
Typically, this conditional probability is factorized as the product of conditional probabilities
of the next word in the sequence:
\begin{eqnarray}
 p(\bm{y}|\bm{x}) = \prod_{t=1}^{M} p(y_t|\bm{y}_{<t}, \bm{x}), \nonumber
\end{eqnarray}
where $\bm{y}_{<t} = (y_1, y_2, y_3, ..., y_{t-1})$.
In the following, we describe how to compute $p(y_t|\bm{y}_{<t}, \bm{x})$.

\subsection{Encoder}

We use the bi-directional RNN (BiRNN) as encoder which has been shown effective
in neural machine translation \cite{bahdanau14} and speech recognition
\cite{schuster97,graves13}.

A BiRNN processes the source sentence for both forward and backward directions
with two separate RNNs.
During the encoding process, the BiRNN computes both forward hidden states $(\encfh{1},
\encfh{2}, ..., \encfh{N})$ and backward hidden states $(\encbh{1}, \encbh{2},
..., \encbh{N})$ as follows:
\begin{eqnarray}
\encfh{t} = g(\encfh{t-1}, x_t), \nonumber \\
\encbh{t} = g(\encbh{t+1}, x_t). \nonumber 
\end{eqnarray}
While $g$ can be any kind of recurrent unit, we use long short-term memory (LSTM) \cite{hochreiter97} networks that have memory cells for both directions ($\encfc{t}$
and $\encbc{t}$).

After encoding, we set the initial hidden states $\dech{0}$ and memory-cell
$\decc{0}$ of the decoder as follows:
\begin{eqnarray}
  \dech{0} &=& \encbh{1}, \nonumber \\
  \decc{0} &=& \encbc{1}. \nonumber
\end{eqnarray}

\subsection{Decoder and Attender}

Our decoder is based on an RNN with LSTM $g$:
\begin{eqnarray}
\dech{t} = g(\dech{t-1}, x_t). \nonumber
\end{eqnarray}

We also use the attention mechanism developed by Luong et al. 
\shortcite{luong15},
which uses $\dech{t}$ to compute contextual information $\bm{d}_t$ of time step $t$.
We first summarize the forward and backward encoder states by taking their sum $\bar{\bm{h}}_i = \encfh{i} + \encbh{i}$, 
and then calculate the context vector $\bm{d}_t$ as the weighted sum of these summarized vectors:
\begin{eqnarray}
 \bm{d}_t = \sum_{i} \bm{a}_{ti} \bar{\bm{h}}_i, \nonumber
\end{eqnarray}
where $\bm{a}_t$ is the weight at the $t$-th step for $\bar{\bm{h}}_i$ computed by a softmax
operation:
%
%softmax operation:
\begin{eqnarray}
a_{ti} &=& \frac{\text{exp}(\dech{t} \cdot \bar{\bm{h}}_i)}{\sum_{\bar{\bm{h}}'}
\text{exp}(\dech{t}\cdot \bar{\bm{h}}')}. \nonumber
\end{eqnarray}

After context vector $\bm{d}_t$ is calculated, the model updates the
distribution over the next word as follows:
\begin{eqnarray}
\tilde{\bm{s}}_t &=& \text{tanh}(W_{hs} [\dech{t}; \bm{d}_t] + \bm{b}_{hs}), \nonumber \\ 
p(y_t|\bm{y}_{<t}, \bm{x}) &=& \text{softmax}(W_{so} \tilde{\bm{s}_t} + \bm{b}_{so}). \nonumber
\end{eqnarray}
Note that $\tilde{\bm{s}}_t$ is also provided as input to the LSTM with $y_t$ for the next step,
which is called the {\it input feeding} architecture \cite{luong15}.

\subsection{Training and Decoding}

The training objective of our models is to maximize log likelihood of the
sentence-summary pairs in a given training set $D$:
% objective
\begin{eqnarray}
  L_t(\theta) = \sum_{(\bm{x},\bm{y}) \in D} \log p(\bm{y}|\bm{x};\theta), \nonumber \\
  p(\bm{y}|\bm{x};\theta) = \prod_{t} p(y_t|\bm{y}_{<t}, \bm{x}). \nonumber
\end{eqnarray}
Once models are trained,  we use beam search to find the output that
maximizes the conditional probability.

\section{Controlling Length in Encoder-decoders}

In this section, we propose our four methods that can control the length of the
output in the encoder-decoder framework.
In the first two methods, the decoding process is used to control the
output length without changing the model itself.
In the other two methods, the model itself has been changed and is trained to
obtain the capability of controlling the length.
Following the evaluation dataset used in our experiments, we use bytes as the
unit of length, although our models can use either words or bytes as necessary.

\subsection{{\modelA}: Beam Search without EOS Tags}
\label{sec:fixlen}

The first method we examine is a decoding approach similar to the one taken in many recent NSS methods that is slightly less ad-hoc.
In this method, we inhibit the decoder from generating the EOS tag by assigning it a score of $-\infty$.
Since the model cannot stop the decoding process by itself, 
we simply stop the decoding process when the length of output sequence reaches the
desired length.
More specifically, during beam search, when the length of the sequence generated
so far exceeds the desired length, the last word is replaced with the EOS tag
and also the score of the last word is replaced with the score of the EOS tag
({\it EOS replacement}).

\subsection{{\modelB}: Discarding Out-of-range Sequences}

Our second decoding method is based on discarding out-of-range sequences, and is not
inhibited from generating the EOS tag, allowing it to decide
when to stop generation.
Instead, we define the legitimate range of the sequence by setting minimum
and maximum lengths.
Specifically, in addition to the normal beam search procedure,
we set two rules:
\begin{itemize}
\item If the model generates the EOS tag when the output sequence is
shorter than the minimum length, we discard the sequence from the beam.
\item If the generated sequence exceeds the maximum length, we also discard the
sequence from the beam.
We then replace its last word with the EOS tag and add this sequence to
the beam ({EOS replacement} in Section \ref{sec:fixlen}).\footnote{This is a workaround to
prevent the situation in which all sequences are discarded from a beam.}
\end{itemize}

In other words, we keep only the sequences that contain the EOS tag and are in
the defined length range.
This method is a compromise that allows the model some flexibility to plan the
generated sequences, but only within a certain acceptable length range.

It should be noted that this method needs a larger beam size if the desired length is very different
from the average summary length in the training data, as it will need to preserve hypotheses that
have the desired length.

\subsection{{\modelC}: Length Embedding as Additional Input for the LSTM}

\begin{figure}[t]
  \centering
  \includegraphics[keepaspectratio,scale=0.37]{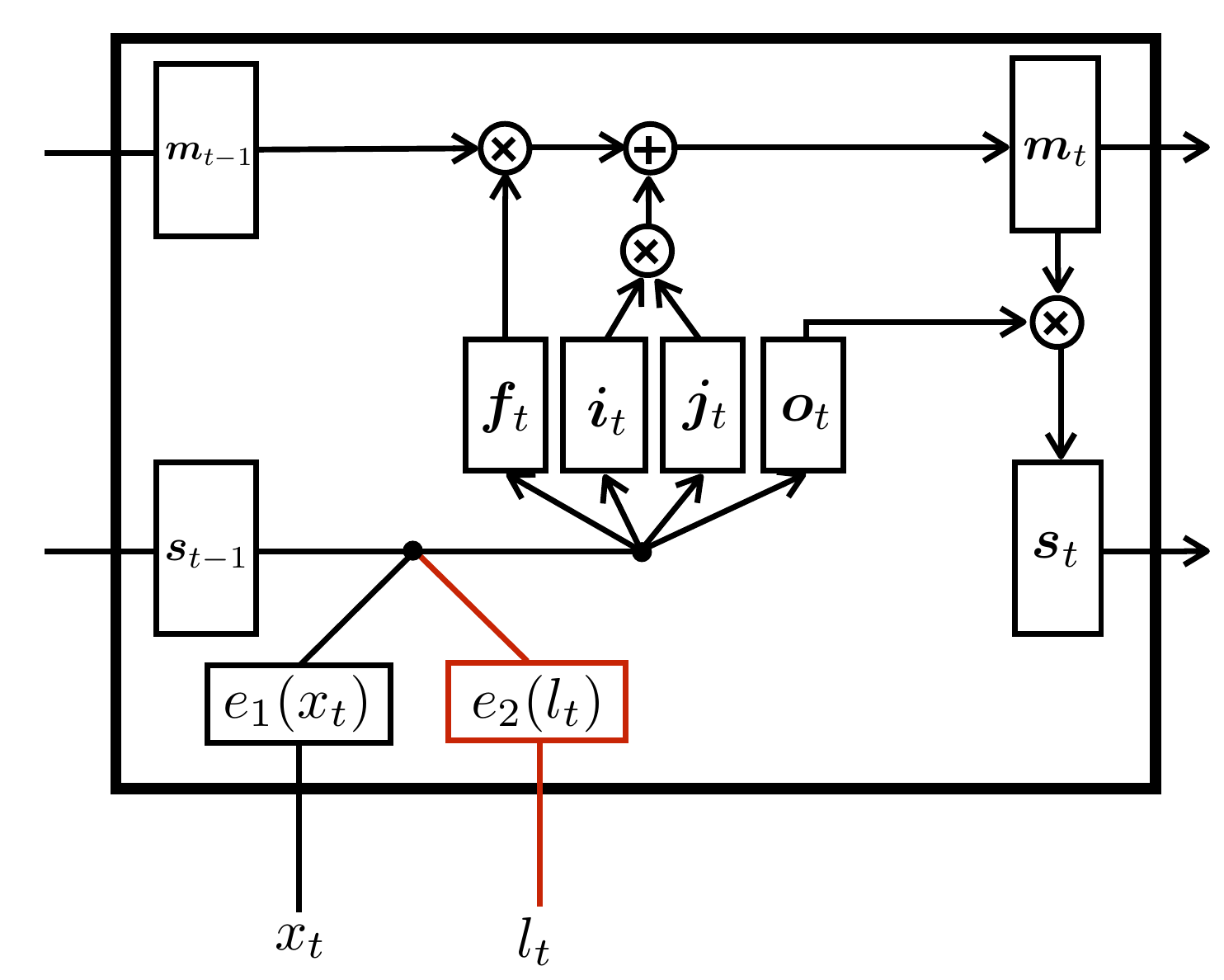}
  \caption{\modelC: {\it remaining length} is used as additional input for
the LSTM of the decoder.}
  \label{fig:lenemb}
\end{figure}

Our third method is a learning-based method specifically trained to control the length of
the output sequence.
Inspired by previous work that has demonstrated that additional inputs to decoder models
can effectively control the characteristics of the output \cite{wen15,li16b}, this model
provides information about the length in the form of an additional input to the net.
Specifically, the model uses an embedding $e_2(l_t) \in
\mathbb{R}^{D}$ for each potential desired length, which is parameterized
by a length embedding matrix $W_{le} \in \mathbb{R}^{D{\times}L}$ where $L$ is
the number of length types.
In the decoding process, we input the embedding of the {\it remaining length} $l_t$ as 
additional input to the LSTM (Figure \ref{fig:lenemb}).
$l_t$ is initialized after the encoding process and updated during the
decoding process as follows:
\begin{eqnarray}
l_{1} &=& length, \nonumber \\
l_{t+1} &=& \left\{ 
\begin{array}{ll} 
  0 & (l_t - byte(y_t) \leq 0) \\
  l_t - byte(y_t) & (\text{otherwise}), \nonumber
\end{array} \right.
\end{eqnarray}
where $byte(y_t)$ is the length of output word $y_t$ and $length$ is the desired
length.
We learn the values of the length embedding matrix $W_{le}$ during training.
This method provides additional information about the amount of length remaining in the output sequence,
allowing the decoder to ``plan'' its output based on the remaining number of words it can generate.
%

%%%%%%%%%%%%%%%%%%%%%%%%%%%%%%%%%%%%%%%%%%%%%%%%%%

%%%%%%%%%%%%%%%%%%%%%%%%%%%%%%%%%%%%%%%%%%%%%%%%%%
\begin{figure}[t]
  \vspace{-5mm}
  \centering
  \includegraphics[keepaspectratio,scale=0.45]{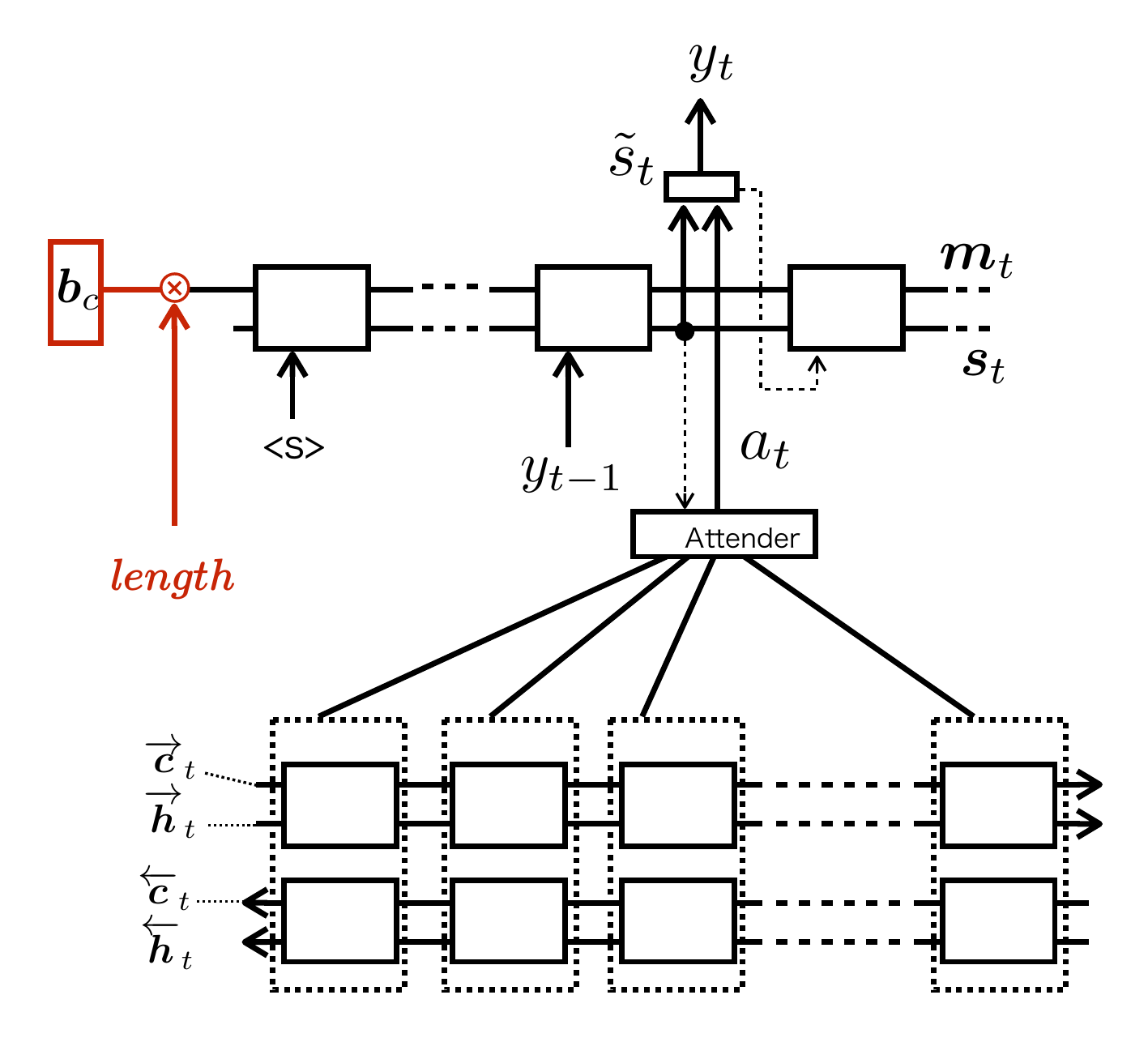}
  \vspace{-4mm}
  \caption{{\modelD}: initial state of the decoder's memory cell $m_0$ manages output
length.}
  \label{fig:leninit}
\end{figure}

\begin{figure*}[t]
      \centering
  \begin{minipage}[b]{0.32\linewidth}
    \centering
    \includegraphics[keepaspectratio, scale=0.4]{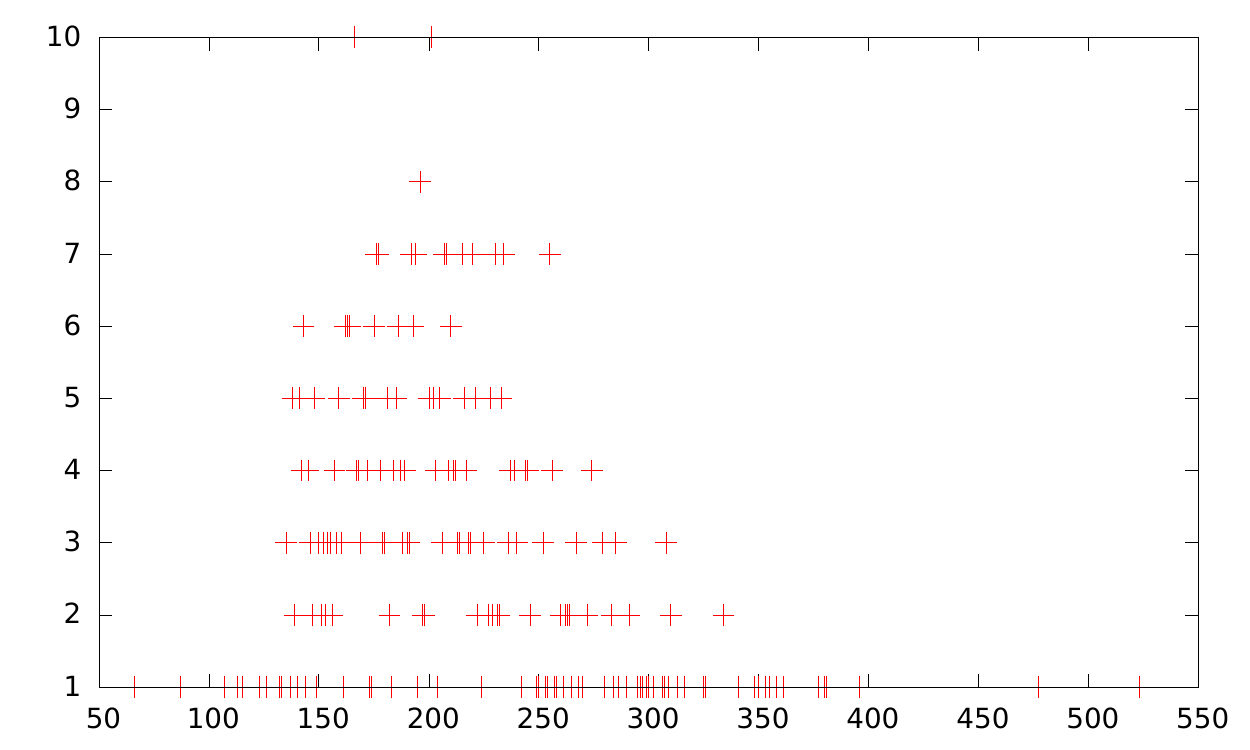}
    \subcaption{first sentence (206.91)}\label{fig:article_hist}
  \end{minipage}
  \begin{minipage}[b]{0.32\linewidth}
    \centering
    \includegraphics[keepaspectratio, scale=0.4]{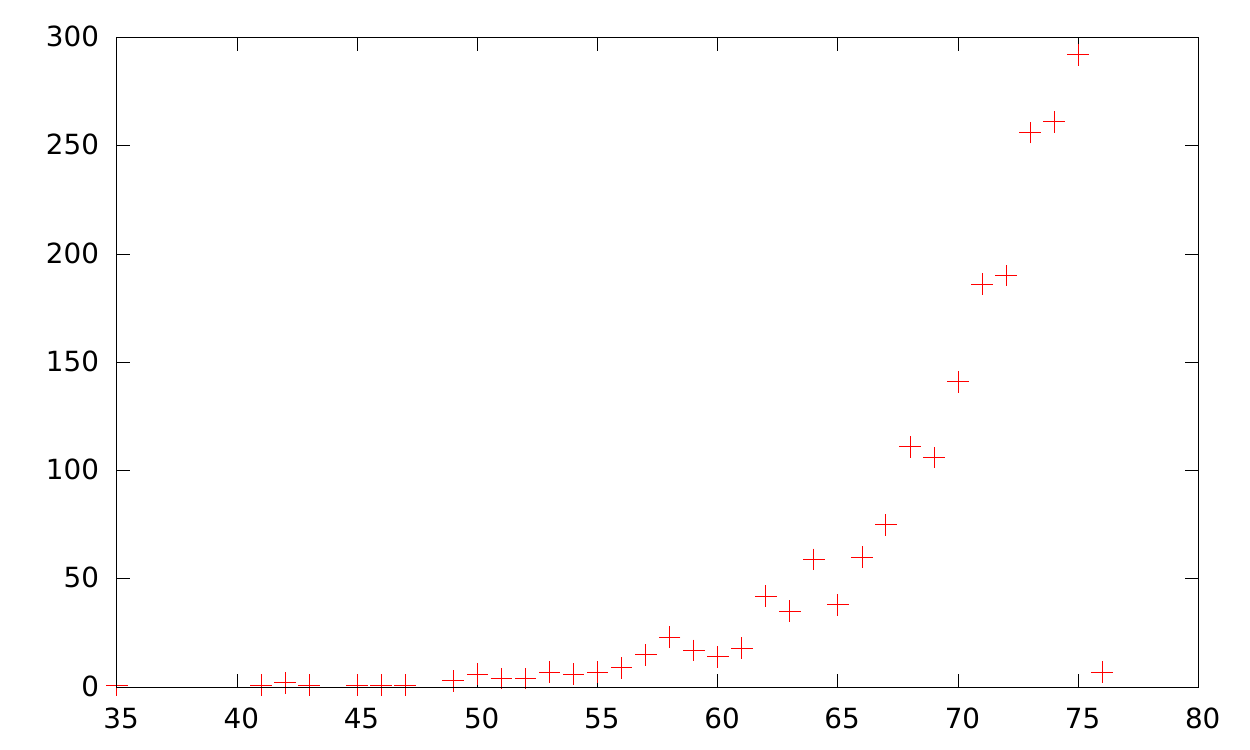}
    \subcaption{summary (70.00)}\label{fig:title_hist}
  \end{minipage}
  \begin{minipage}[b]{0.32\linewidth}
    \centering
    \includegraphics[keepaspectratio, scale=0.4]{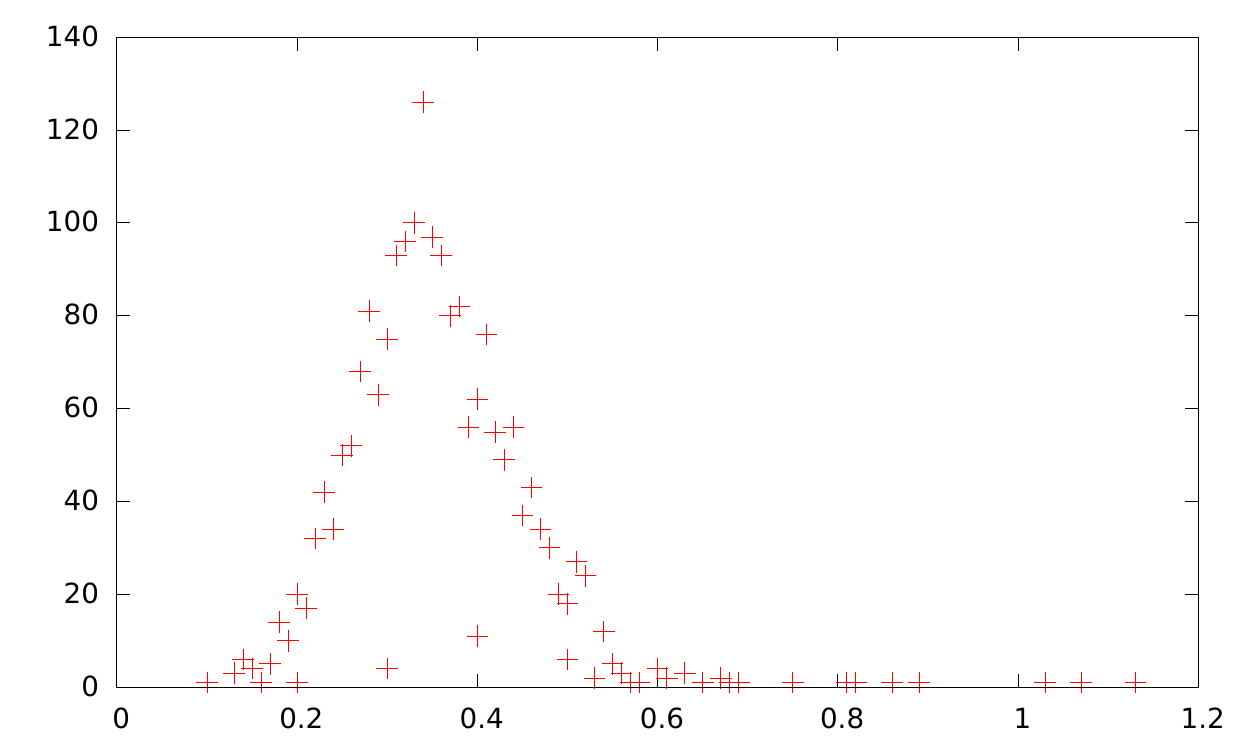}
    \subcaption{ratio (0.35)}\label{fig:ratio_hist}
  \end{minipage}
  \caption{Histograms of first sentence length, summary length, and their ratio in
DUC2004.}
  \label{fig:duc_hist}
\end{figure*}

\subsection{\modelD: Length-based Memory Cell Initialization}

While {\modelC} inputs the {\it remaining length} $l_t$ to the decoder at each step of
the decoding process, the {\modelD} method inputs the desired length once at the initial
state of the decoder. Figure \ref{fig:leninit} shows the architecture of {\modelD}.
Specifically, the model uses the memory cell $\decc{t}$ to control the output length
by initializing the states of decoder (hidden state
$\dech{0}$ and memory cell $\decc{0}$) as follows:
{ \small
\begin{eqnarray}
\dech{0} &=& \encbh{1}, \nonumber \\
\decc{0} &=& \bm{b}_{c} * length, \label{eqn:leninit}
\end{eqnarray}
}
where $\bm{b}_c \in \mathbb R^{H}$ is a trainable parameter and $length$ is the
desired length.

While the model of {\modelC} is guided towards the appropriate output length by inputting
the remaining length at each step, this {\modelD} attempts to provide the model with the
ability to manage the output length on its own using its inner state.
Specifically, the memory cell of LSTM networks is suitable for this endeavour, as it
is possible for LSTMs to learn functions that, for example, subtract a fixed amount from
a particular memory cell every time they output a word.
Although other ways for managing the length are also possible,\footnote{For example,
we can also add another memory cell for managing the length.} we found this 
approach to be both simple and effective.

\section{Experiment}

\subsection{Dataset}

We trained our models on a part of the Annotated English Gigaword corpus \cite{agiga},
which Rush et al. \shortcite{rush15} constructed for sentence summarization.
We perform preprocessing using the standard script for the
dataset\footnote{https://github.com/facebook/NAMAS}.
The dataset consists of approximately 3.6 million pairs of the first sentence from each
source document and its headline.
Figure \ref{fig:agiga_hist} shows the length histograms of the summaries in the
training set.
The vocabulary size is 116,875 for the source documents and 67,564 for the target
summaries including the beginning-of-sentence, end-of-sentence, and
unknown word tags.
For {\modelC} and {\modelD}, we input the length of each headline during training.
Note that we do not train multiple summarization models for each headline length, but
a single model that is capable of controlling the length of its output.

We evaluate the methods on the evaluation set of DUC2004 task-1 (generating very
short single-document summaries).
In this task, summarization systems are required to create a very short summary
for each given document.
Summaries over the length limit (75 bytes) will be truncated and there is no
bonus for creating a shorter summary. 
The evaluation set consists of 500 source documents and 4 human-written
(reference) summaries for each source document.
Figure \ref{fig:duc_hist} shows the length histograms of the summaries in the
evaluation set.
Note that the human-written summaries are not always as long as 75 bytes.
We used three variants of ROUGE \cite{rouge} as evaluation metrics:
ROUGE-1 (unigram), ROUGE-2 (bigram), and ROUGE-L (longest common subsequence).
The two-sided permutation test \cite{chinchor92} was used for statistical
significance testing ($p \leq 0.05$).

\subsection{Implementation}

We use Adam \cite{adam} ($\alpha$=0.001, $\beta_1$=0.9, $\beta_2$=0.999,
$eps$=$10^{-8}$)
to optimize parameters with a mini-batch of size 80.
Before every 10,000 updates, we first sampled 
800,000 training examples and made groups of 80 examples with the same source
sentence length, and shuffled the 10,000 groups.

We set the dimension of word embeddings to 100 and that of the hidden state to 200.
For LSTMs, we initialize the bias of the forget gate to 1.0 and use 0.0 for the other gate biases \cite{jozefowicz15}.
We use Chainer \cite{chainer} to implement our models.
For {\modelC}, we set $L$ to 300, which is larger than the longest summary lengths in our
dataset (see Figure \ref{fig:agiga_hist}-(b) and Figure \ref{fig:duc_hist}-(b)).

For all methods except {\modelB}, we found a beam size of 10 to be sufficient,
but for {\modelB} we used a beam size of 30 because it more
aggressively discards candidate sequences from its beams during decoding.

% submit version
% \begin{table*}[t]
%   \centering
%   %\scriptsize
%   %\tabcolsep=3mm
%   \begin{tabular}{lccccccccc} \bhline{0.8pt}
%    & \multicolumn{3}{c}{30 byte} & \multicolumn{3}{c}{50 byte} & \multicolumn{3}{c}{75 byte}\\ 
%   model   & R-1        & R-2       & R-L        & R-1        & R-2       & R-L        & R-1        & R-2       & R-L\\ \hline
%   \modelA & \bf{14.35} & 3.10      & \bf{13.24} & 20.01$^*$      & 5.99      & 18.27$^*$      & 25.88$^*$      & 7.93      & 23.07$^*$ \\
%   \modelB & 14.32      & 3.13      & 13.23      & 20.08$^*$      & 5.75      & 18.19$^*$      & 26.02      & 7.69$^*$      & 22.78$^*$ \\
%   \modelC & 14.24      & 3.22      & 13.02      & 20.79      & 5.98      & 18.57 & \bf{26.73} & \bf{8.40} & \bf{23.88} \\ 
%   \modelD & 14.31      & \bf{3.28} & 13.20      & \bf{20.88} & \bf{6.17} &
% \bf{19.01} & 25.87$^*$      & 8.28      & 23.25 \\ \bhline{0.8pt}
%   \end{tabular}
%   \caption{ROUGE scores with various length limits.  The scores with $*$ are
% significantly worse than the best score in the column (bolded).}
%   \label{tab:rouges}
% \end{table*}

% camera ready version (with backoff)
\begin{table*}[t]
  \centering
  %\scriptsize
  \tabcolsep=2.5mm
  \begin{tabular}{lccccccccc} \bhline{0.8pt}
    & \multicolumn{3}{c}{30 byte} & \multicolumn{3}{c}{50 byte} & \multicolumn{3}{c}{75 byte}\\ 
  model   & R-1        & R-2       & R-L        & R-1        & R-2       & R-L        & R-1        & R-2       & R-L\\ \hline
  % \modelA & \bf{14.34} & 3.10      & {\bf 13.23}      & 20.00$^*$  & 5.98 & 18.26$^*$ & 25.87$^*$  & 7.93      & 23.07$^*$ \\
  % \modelB & 13.83      & 3.08      & 12.88 & 20.08$^*$  & 5.74            & 18.19$^*$ & 26.01      & 7.69$^*$  & 22.77$^*$ \\
  % \modelC & 14.23      & 3.21      & 13.02      & 20.78      & 5.97       & 18.57     & \bf{26.73} & 8.39 & 23.88 \\ 
  % \modelD & 14.31      & 3.27 & 13.19      & \bf{20.87} & 6.16            & 19.00     & 25.87$^*$  & 8.27      & 23.24 \\ \bhline{0.8pt}
  % 
  %%\modelA & \bf{14.34} & 3.10      & \bf{13.23} & 20.00$^*$  & 5.98      & 18.26$^*$  & 25.87$^*$  & 7.93      & 23.07$^*$  \\
  %%\modelB & 13.83$^*$  & 3.08      & 12.88      & 20.08$^*$  & 5.74      & 18.19$^*$  & 26.01      & 7.69$^*$  & 22.77$^*$  \\
  %%\modelC & 14.23      & 3.21      & 13.02      & 20.78      & 5.97      & 18.57      & \bf{26.73} & \bf{8.39} & \bf{23.88} \\
  %%\modelD & 14.31      & \bf{3.27} & 13.19      & \bf{20.87} & \bf{6.16} & \bf{19.00} & 25.87$^*$  & 8.27      & 23.24      \\ \hline
  %%\modelC$_{(0, \infty)}$ & 13.75 & 3.30 & 12.68 & 20.62 & {\bf 6.22} & 18.64 & 26.42 & 8.26 & 23.59 \\  
  %%\modelD$_{(0, \infty)}$ & 13.92 & {\bf 3.49} & 12.90 & 20.87 & 6.19 & {\bf 19.09} & 25.29 & 8.00 & 22.71 \\ \bhline{0.8pt}
  {\modelA}                 & \bf{14.34} & 3.10$^*$  & \bf{13.23} & 20.00$^*$    & 5.98      & 18.26$^*$  & 25.87$^*$  & 7.93      & 23.07$^*$ \\
  {\modelB}                 & 13.83$^*$  & 3.08$^*$  & 12.88      & 20.08$^*$    & 5.74      & 18.19$^*$  & 26.01      & 7.69$^*$  & 22.77$^*$ \\
  {\modelC}$_{(0, L)}$        & 14.23      & 3.21      & 13.02      & 20.78        & 5.97      & 18.57      & \bf{26.73} & \bf{8.39} & \bf{23.88} \\
  {\modelD}$_{(0, L)}$        & 14.31      & 3.27      & 13.19      & \bf{20.87}   & 6.16      & 19.00      & 25.87      & 8.27      & 23.24  \\ \hline
  {\modelC}$_{(0, \infty)}$ & 13.75      & 3.30      & 12.68      & 20.62        & \bf{6.22} & 18.64      & 26.42      & 8.26      & 23.59  \\
  {\modelD}$_{(0, \infty)}$ & 13.92      & \bf{3.49} & 12.90      & 20.87        & 6.19      & \bf{19.09} & 25.29$^*$  & 8.00      & 22.71$^*$ \\ \hline
  %
  %\modelC$_{(N-5, N)}$    & 14.23 & 3.21 & 13.02 & 20.84 & 6.00 & 18.61 & 26.75
 %{\bf 8.42} & {\bf 23.89} \\ 
  %\modelD$_{(N-5, N)}$    & 14.30 & 3.26 & 13.19 & 20.91 & 6.17 & 19.01 & 26.00 & 8.27 & 23.31 \\ \bhline{0.8pt}
  %\modelC$_{(0, \infty)}$ & 13.75 & 3.30 & 12.68 & 20.62 & {\bf 6.22} & 18.64 & 26.42 & 8.26 & 23.59 \\ 
  %\modelD$_{(0, \infty)}$ & 13.92 & {\bf 3.49} & 12.90 & 20.87 & 6.19 & {\bf 19.09} & 25.29 & 8.00 & 22.71 \\ \bhline{0.8pt}
  %
  \end{tabular}
  \caption{ROUGE scores with various length limits.  The scores with $*$ are
significantly worse than the best score in the column (bolded).}
  \label{tab:rouges}
\end{table*}

\begin{table*}[t]
  %\scriptsize
  \footnotesize
  \center
  \begin{tabular}{rp{0.75\linewidth}} \bhline{0.8pt}
source{\ \ }  & five-time world champion michelle kwan withdrew from the \#\#\#\# us figure
skating championships on wednesday , but will petition us skating officials for
the chance to compete at the \#\#\#\# turin olympics . \\ \hline
reference{\ \ }  & injury leaves kwan 's olympic hopes in limbo \\ \hline \hline
{\modelA} (30){\ \ } & kwan withdraws from us gp \\
          (50){\ \ } & kwan withdraws from us skating championships \\
          (75){\ \ } & kwan pulls out of us figure skating championships for turin olympics \\ \hline
{\modelB} (30){\ \ } & kwan withdraws from us gp \\
          (50){\ \ } & kwan withdraws from figure skating championships \\
          (75){\ \ } & kwan pulls out of us figure skating championships for turin olympics bid \\ \hline
{\modelC} (30){\ \ } & kwan withdraws from us skating \\
          (50){\ \ } & kwan withdraws from us figure skating championships \\
          (75){\ \ } & world champion kwan withdraws from \#\#\#\# olympic figure skating championships \\ \hline
{\modelD} (30){\ \ } & kwan quits us figure skating \\
          (50){\ \ } & kwan withdraws from \#\#\#\# us figure skating worlds \\
          (75){\ \ } & kwan withdraws from \#\#\#\# us figure skating
championships for \#\#\#\# olympics \\ \bhline{0.8pt}
  \end{tabular}
  \caption{Examples of the output of each method with various specified lengths.}
  \label{tab:example1}
\end{table*}

\begin{table*}[t]
  %\scriptsize
  \footnotesize
  \center
  \begin{tabular}{rp{0.75\linewidth}} \bhline{0.8pt}
  source{\ \ }  & at least two people have tested positive for the bird flu virus in
  eastern turkey , health minister recep akdag told a news conference wednesday
.  \\ \hline
  reference{\ \ }  & two test positive for bird flu virus in turkey \\
  \hline \hline
  %{\modelA} (30){\ \ } & two people tested positive for \\
  %          (50){\ \ } & two people tested positive for bird flu in eastern \\
  %          (75){\ \ } & two people tested positive for bird flu in eastern turkey says health minister \\ 
  {\modelA} (30){\ \ } & two infected with bird flu \\
            (50){\ \ } & two infected with bird flu in eastern turkey \\
            (75){\ \ } & two people tested positive for bird flu in eastern turkey says minister\\
  \hline
  %{\modelB} (30){\ \ } & two infected in bird flu case \\
  %          (50){\ \ } & two tests positive for bird flu in eastern turkey \\
  %          (75){\ \ } & two people tested positive for bird flu in eastern turkey says minister \\
  {\modelB} (30){\ \ } &  two infected with bird flu \\ 
            (50){\ \ } &  two more infected with bird flu in eastern turkey\\ 
            (75){\ \ } &  two people tested positive for bird flu in eastern turkey says minister\\ 
  \hline
  {\modelC} (30){\ \ } & two bird flu cases in turkey \\
            (50){\ \ } & two confirmed positive for bird flu in eastern turkey \\
            (75){\ \ } & at least two bird flu patients test positive for bird flu in eastern turkey \\
  \hline
  {\modelD} (30){\ \ } & two cases of bird flu in turkey \\
            (50){\ \ } & two people tested positive for bird flu in turkey \\
            (75){\ \ } & two people tested positive for bird flu in eastern turkey health conference \\
  \bhline{0.8pt}
  \end{tabular}
  \caption{More examples of the output of each method.}
  \label{tab:example2}
\end{table*}

\section{Result}

\subsection{ROUGE Evaluation}

Table \ref{tab:rouges} shows the ROUGE scores of each method with various length
limits (30, 50 and 75 byte).
Regardless of the length limit set for the summarization methods,
we use the same reference summaries.
Note that, {\modelA} and {\modelB} generate the summaries with a hard
constraint due to their decoding process, which allows them to follow the hard constraint on length.
Hence, when we calculate the scores of {\modelC} and {\modelD}, we 
impose a hard constraint on length to make the comparison fair (i.e. {\modelC$_{(0,
L)}$} and {\modelD$_{(0, L)}$} in the table).
Specifically, we use the same beam search as that for {\modelB} with minimum length of 0.

For the purpose of showing the length control capability of {\modelC} and {\modelD},
we show at the bottom two lines the results of the standard beam search without
the hard constraints on the length\footnote{ {\modelB} is equivalence to the
standard beam search when we set the range as $(0, \infty)$.}.
We will use the results of {\modelC$_{(0, \infty)}$} and {\modelD$_{(0, \infty)}$}
in the discussions in Sections \ref{sec:examples} and \ref{sec:lencon}.

The results show that the learning-based methods ({\modelC} and {\modelD}) tend
to outperform decoding-based methods ({\modelA} and {\modelB}) for the longer summaries
of 50 and 75 bytes.
However, in the 30-byte setting, there is no significant difference between
these two types of methods.
We hypothesize that this is because average compression rate in the training data is 30\% (Figure
\ref{fig:agiga_hist}-(c)) while the 30-byte setting forces the model to generate
summaries with 15.38\% in average compression rate, and thus the learning-based models
did not have enough training data to learn compression at such a steep rate.

\subsection{Examples of Generated Summaries}
\label{sec:examples}

Tables \ref{tab:example1} and \ref{tab:example2} show examples from the
validation set of the Annotated Gigaword Corpus.
The tables show that all models, including both learning-based methods and
decoding-based methods, can often generate well-formed sentences. 

We can see various paraphrases of ``\#\#\#\# us figure
championships''\footnote{Note that ``\#'' is a normalized number and ``us'' is ``US'' (United
States).} and ``withdrew''.
Some examples are generated as a single noun phrase ({\modelC}(30) and
{\modelD}(30)) which may be suitable for the short length setting.

\begin{table*}[t]
  \begin{minipage}[b]{0.37\linewidth}
    \centering
    \scriptsize
    \begin{tabular}{ccl} \bhline{0.8pt} \hline
    log$p(\bm{y}|\bm{x})$ & byte & candidate summary     \\ \hline
    -4.27 & 31 & two cases of bird flu in turkey  \\
    -4.41 & 28 & two bird flu cases in turkey     \\
    -4.65 & 30 & two people tested for bird flu   \\
    -5.25 & 30 & two people tested in e. turkey   \\
    -5.27 & 31 & two bird flu cases in e. turkey  \\
    -5.51 & 29 & two bird flu cases in eastern    \\
    -5.55 & 32 & two people tested in east turkey \\
    -5.72 & 30 & two bird flu cases in turkey :   \\
    -6.04 & 30 & two people fail bird flu virus   \\
    \bhline{0.8pt}
    \end{tabular}
    \subcaption{the beam of {\modelD}}
    \label{tab:beam_leninit}
  \end{minipage}
  \begin{minipage}[b]{0.65\linewidth}
    \centering
    \scriptsize
    \begin{tabular}{ccl} \bhline{0.8pt} \hline
    log$p(\bm{y}|\bm{x})$ & byte & candidate summary     \\ \hline
    -5.05 & 57 & two people tested positive for bird flu in eastern turkey \\
    -5.13 & 50 & two tested positive for bird flu in eastern turkey \\
    -5.30 & 39 & two people tested positive for bird flu \\
    -5.49 & 51 & two people infected with bird flu in eastern turkey \\
    -5.52 & 32 & two tested positive for bird flu \\
    -5.55 & 44 & two infected with bird flu in eastern turkey \\
    -6.00 & 49 & two more infected with bird flu in eastern turkey \\
    -6.04 & 54 & two more confirmed cases of bird flu in eastern turkey \\
    -6.50 & 49 & two people tested positive for bird flu in turkey \\
    \bhline{0.8pt}
    \end{tabular}
    \subcaption{the beam of the standard encoder-decoder}
    \label{tab:beam_encdec}
  \end{minipage}
  \caption{\protect Final state of the beam when the learning-based model is
instructed to output a 30 byte summary for the source document in Table \ref{tab:example2}.}
  \label{tab:final_beam}
\end{table*}

\subsection{Length Control Capability of Learning-based Models}
\label{sec:lencon}

\begin{figure*}[t]
  \begin{minipage}[b]{0.32\linewidth}
    \centering
    \includegraphics[keepaspectratio, scale=0.295]{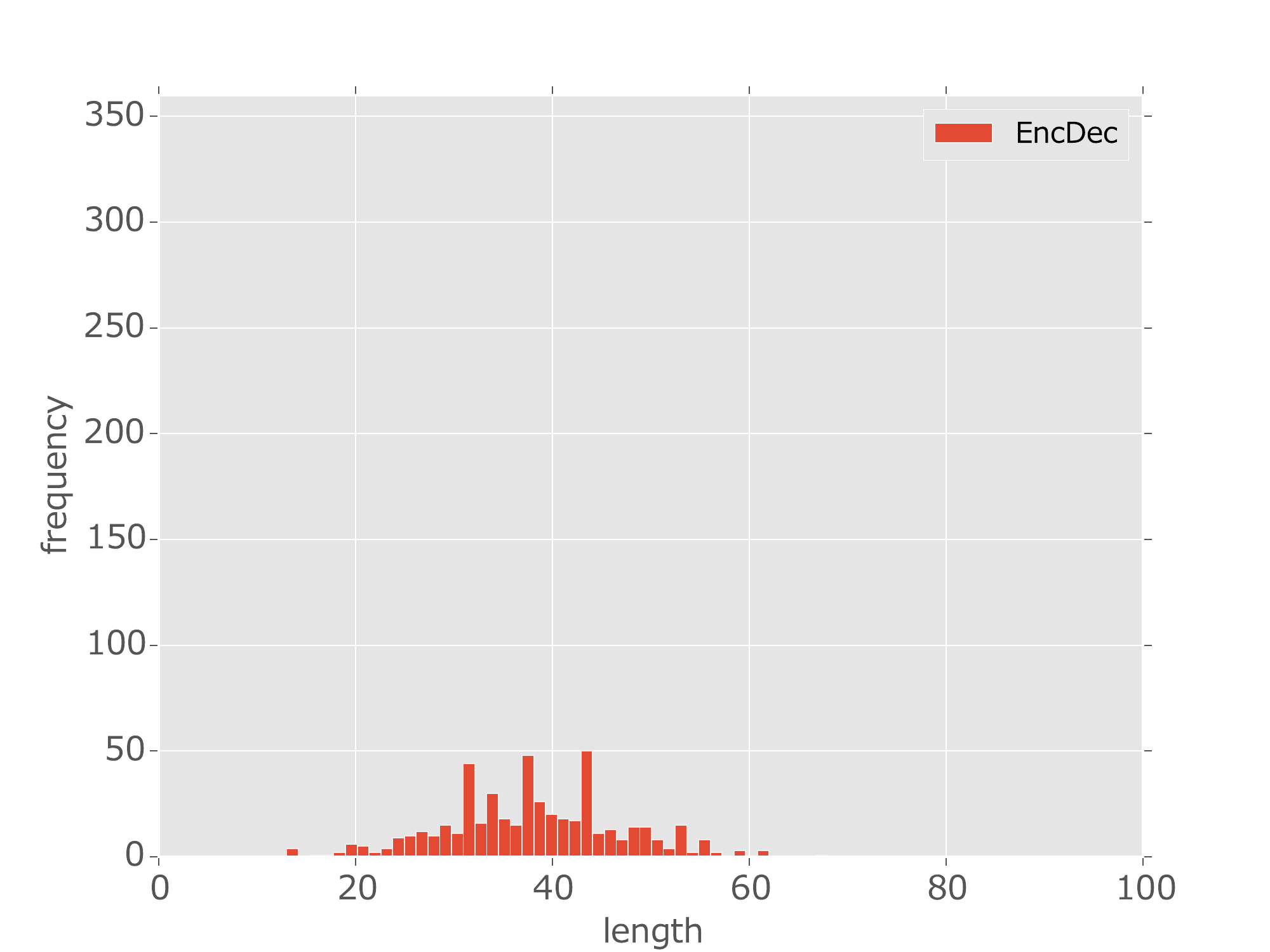}
    \subcaption{\footnotesize encoder-decoder}\label{fig:hist_encdec}
  \end{minipage}
  \begin{minipage}[b]{0.32\linewidth}
    \centering
    \includegraphics[keepaspectratio, scale=0.295]{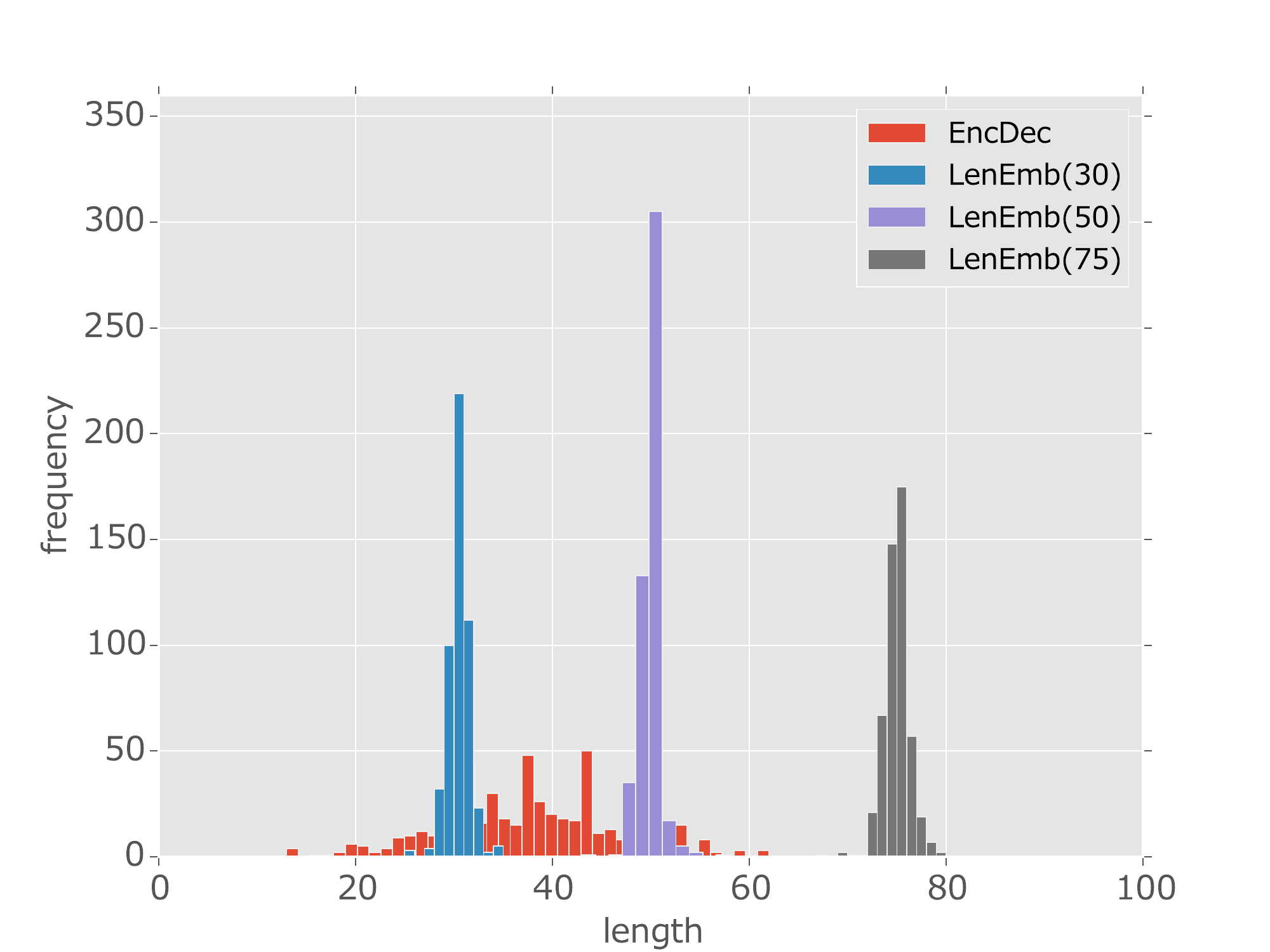}
    \subcaption{\footnotesize \modelC}\label{fig:hist_lenemb}
  \end{minipage}
  \begin{minipage}[b]{0.32\linewidth}
    \centering
    \includegraphics[keepaspectratio, scale=0.295]{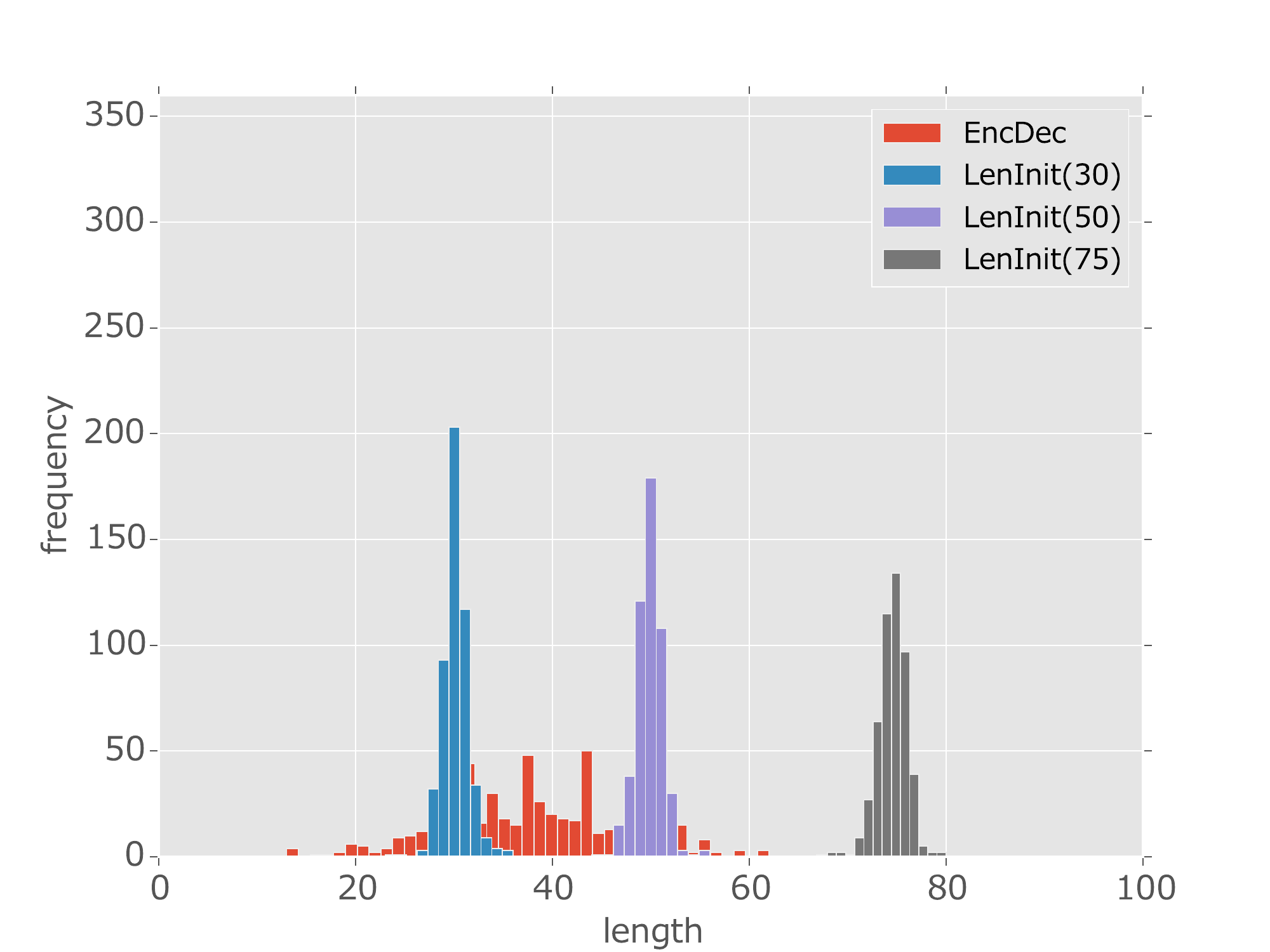}
    \subcaption{\footnotesize \modelD}\label{fig:hist_leninit}
  \end{minipage}
  \caption{Histograms of output lengths generated by (a) the standard encoder-decoder
, (b) {\modelC}, and (c) {\modelD}. 
For {\modelC} and {\modelD}, the bracketed numbers in each region are the
desired lengths we set.}

\label{fig:length_hist}
\end{figure*}

%first shitogram, next one beam
Figure \ref{fig:length_hist} shows histograms of output length
from the standard encoder-decoder,
%\footnote{
%{\modelA} and {\modelB} use this model with specific decoding to control the
%output lengths})
{\modelC}, and {\modelD}.
While the output lengths from the standard model disperse widely,
the lengths from our learning-based models are concentrated to the desired length.
These histograms clearly show the length controlling capability of our 
learning-based models.

Table \ref{tab:final_beam}-(a) shows the final state of the beam when {\modelD}
generates the sentence with a length of 30 bytes for the example 
with standard beam search in Table \ref{tab:example2}.
We can see all the sentences in the beam are generated with length close to
the desired length.
This shows that our method has obtained the ability to control the output length
as expected.
For comparison, Table \ref{tab:final_beam}-(b) shows the final state of the beam
if we perform standard beam search in the standard encoder-decoder model (used in {\modelA} and {\modelB}).
Although each sentence is well-formed, the lengths of them are much more varied.

\subsection{Comparison with Existing Methods}

Finally, we compare our methods to existing methods on standard settings of
the DUC2004 shared task-1.
Although the objective of this paper is not to obtain state-of-the-art scores on
this evaluation set, it is of interest whether our length-controllable models are
competitive on this task.
Table \ref{tab:rouges_compare} shows that the scores of our methods, which are
copied from Table \ref{tab:rouges}, in addition to the scores of some existing
methods. 
ABS \cite{rush15} is the most standard model of neural sentence summarization
and is the most similar method to our baseline setting ({\modelA}).
This table shows that the score of {\modelA}
is comparable to those of the existing methods. 
The table also shows the {\modelC} and the {\modelD}  have the capability of
controlling the length without decreasing the ROUGE score.

\section{Conclusion}

In this paper, we presented the first examination of the problem of controlling length in
neural encoder-decoder models, from the point of view of summarization.
We examined methods for controlling length of output sequences: two
decoding-based methods ({\modelA} and {\modelB}) and two learning-based methods
({\modelC} and {\modelD}).
The results showed that learning-based methods generally outperform the decoding-based methods,
and the learning-based methods obtained the capability of controlling the output
length without losing ROUGE score compared to existing summarization methods.

\begin{table}[t]
  \centering
  \small
  \begin{tabular}{lccc} \bhline{0.8pt}
  model                                & R-1        & R-2       & R-L \\ \hline
  \modelA                              & 25.88      & 7.93      & 23.07 \\
  \modelB                              & 26.02      & 7.69      & 22.78 \\
  \modelC                              & 26.73      & 8.40      & 23.88 \\ 
  \modelD                              & 25.87      & 8.28      & 23.25 \\ \hline
  ABS$_{\text{\cite{rush15}}}$         & 26.55      & 7.06      & 22.05 \\
  ABS+$_{\text{\cite{rush15}}}$        & 28.18      & \bf{8.49} & 23.81 \\ 
  RAS-Elman$_{\text{\cite{chopra16}}}$ & \bf{28.97} & 8.26      & \bf{24.06} \\
  RAS-LSTM$_{\text{\cite{chopra16}}}$ & 27.41       & 7.69      & 23.06 \\ \bhline{0.8pt}
  \end{tabular}
  \caption{Comparison with existing studies for DUC2004. Note that top four rows
are reproduced from Table \ref{tab:rouges}.}
  \label{tab:rouges_compare}
\end{table}

\section*{Acknowledgments}

This work was supported by JSPS KAKENHI Grant Number JP26280080.
We are grateful to have the opportunity to use the Kurisu server of
Dwango Co., Ltd. for our experiments.

\bibliographystyle{emnlp2016}
\bibliography{reference}

\end{document}